\documentclass[acmsmall]{acmart}

\usepackage{algorithm}
\usepackage{algpseudocode}
\usepackage{multicol}
\usepackage{multirow}
\usepackage{makecell}

\AtBeginDocument{%
  }

\setcopyright{acmlicensed}
\acmJournal{TOMM}
\acmYear{2024}
\acmDOI{10.1145/3706061}

\acmISBN{978-1-4503-XXXX-X/18/06}

\begin{document}


\title{AED-PADA:Improving Generalizability of Adversarial Example Detection via Principal Adversarial Domain Adaptation}

\author{Heqi Peng}
\email{penghq@buaa.edu.cn}
\affiliation{%
  \institution{Beihang University}
  \state{Beijing}
  \country{China}}

\author{Yunhong Wang}
\email{yhwang@buaa.edu.cn}
\affiliation{%
  \institution{Beihang University}
  \state{Beijing}
  \country{China}}

\author{Ruijie Yang}
\email{rjyang@buaa.edu.cn}
\affiliation{%
  \institution{Beihang University}
  \state{Beijing}
  \country{China}}

\author{Beichen Li}
\email{libeichen@buaa.edu.cn}
\affiliation{%
  \institution{Beihang University}
  \state{Beijing}
  \country{China}}

\author{Rui Wang}
\email{wangrui@iie.ac.cn}
\affiliation{%
  \institution{Institute of Information Engineering, Chinese Academy of Sciences}
  \state{Beijing}
  \country{China}}

\author{Yuanfang Guo}
\authornote{Corresponding author.}
\email{andyguo@buaa.edu.cn}
\affiliation{%
  \institution{Beihang University}
  \state{Beijing}
  \country{China}}

\renewcommand{\shortauthors}{Peng et al.}

\begin{abstract}
 Adversarial example detection, which can be conveniently applied in many scenarios, is important in the area of adversarial defense. Unfortunately, existing detection methods suffer from poor generalization performance, because their training process usually relies on the examples generated from a single known adversarial attack and there exists a large discrepancy between the training and unseen testing adversarial examples. To address this issue, we propose a novel method, named Adversarial Example Detection via Principal Adversarial Domain Adaptation (AED-PADA). Specifically, our approach identifies the Principal Adversarial Domains (PADs), i.e., a combination of features of the adversarial examples generated by different attacks, which possesses a large portion of the entire adversarial feature space. Subsequently, we pioneer to exploit Multi-source Unsupervised Domain Adaptation in adversarial example detection, with PADs as the source domains. Experimental results demonstrate the superior generalization ability of our proposed AED-PADA. Note that this superiority is particularly achieved in challenging scenarios characterized by employing the minimal magnitude constraint for the perturbations.
\end{abstract}




\maketitle

\section{Introduction}
Recently, Deep Neural Networks (DNNs) have been playing prominent roles in many applications. Unfortunately, considerable studies have demonstrated that DNNs can be easily deceived if certain imperceptible perturbations are introduced to their inputs~\cite{SzegedyZSBEGF13,NguyenYC15,shi2021,SSA,ILA-DA,a2sc}. These perturbed inputs, also known as adversarial examples, have enforced DNNs to produce erroneous decision and become a significant security concern in safety sensitive scenarios, such as autonomous driving~\cite{attack_driving} and medical diagnosis~\cite{attack_medical}.

Nowadays, adversarial training has been proved to be an effective adversarial defense strategy~\cite{FGSM,AD_tripletloss,XieTGWYL20,adversarial_training_tomm}. However, it requires to possess enough knowledge about the classification models and necessitates substantial computational costs to retrain the classification models. On the contrary, adversarial example detection, a.k.a. adversarial detection, defends against adversarial attacks by distinguishing whether the inputs are benign or manipulated. The mechanism of this type of methods can be efficiently deploy to defend many applications without the requirement of extra knowledge about the core model.

Generalization ability is vital for adversarial detection methods in real-world scenarios, because these methods tend to encounter unseen attacks and they are desired to perform consistently. Current detection techniques~\cite{lid,md,steg,SID,txt_advdetection} typically achieve considerable generalization ability over a few conventional attacks, such as BIM~\cite{BIM}, PGD~\cite{pgd}, and C\&W~\cite{CW}. However, we empirically observe that these methods exhibit instability and inadequate performance against recent attacks, such as SSA~\cite{SSA}, Jitter~\cite{jitter} and ILA-DA~\cite{ILA-DA}. These methods tend to give unsatisfactory generalization performance, because their training process usually relies on a single known adversarial attack and they have not been developed from the perspective of boosting generalization ability.

\begin{figure}[!ht]
\centering
\includegraphics[width=0.70\linewidth]{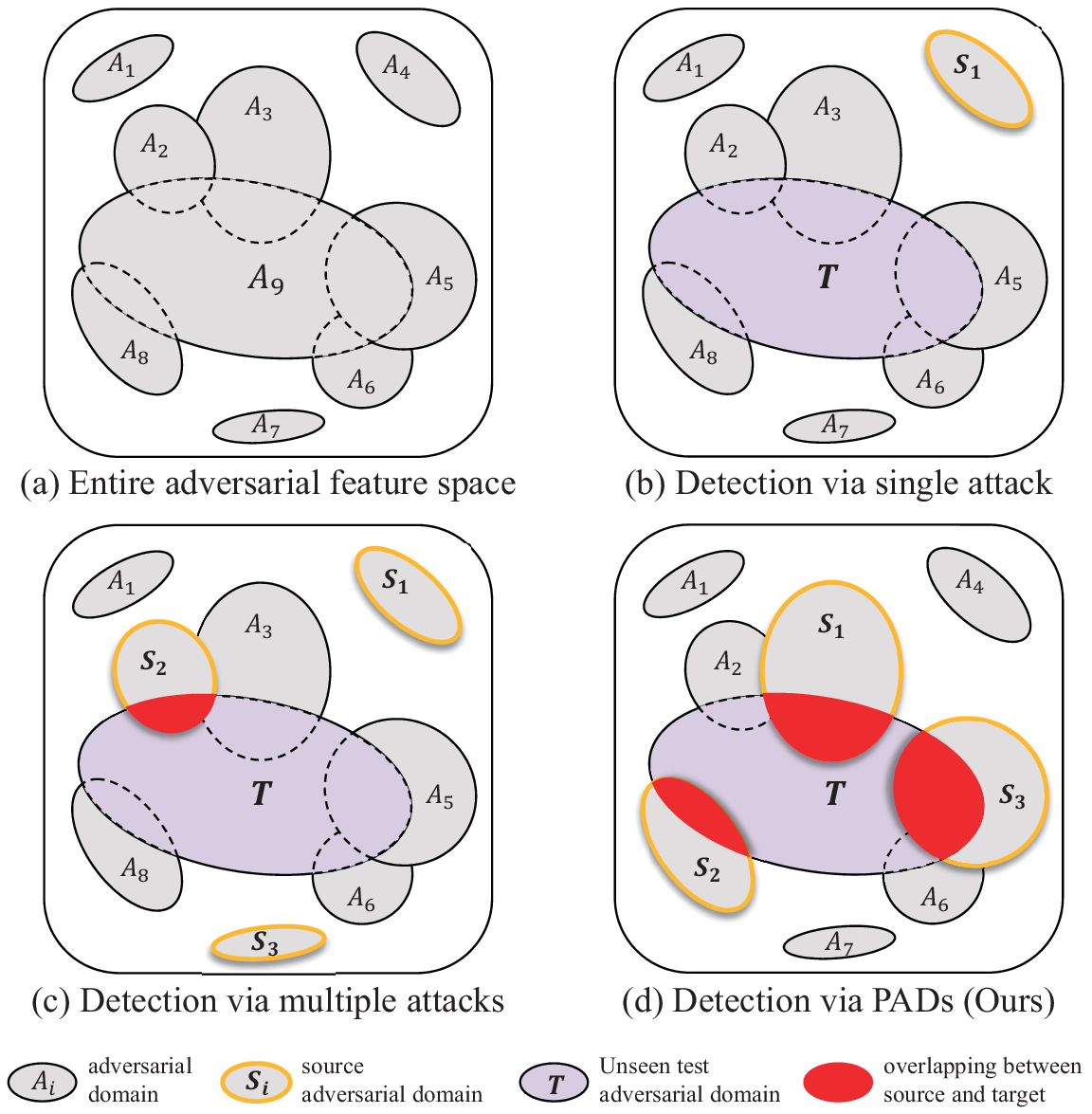} 
\caption{Schematic illustration of our proposed work. (a) represents the entire adversarial feature space as an instance, which contains 9 adversarial domains $\{A_1, A_2,\ldots,A_9\}$. (b) represents the mechanism of existing detection methods, which usually performs the training via a single source domain to detect the examples in the unseen adversarial domain, e.g. $T~(T=A_9)$. (c) is a straightforward solution to improve generalization ability via randomly selecting multiple source domains. (d) presents the intuition behind our work. We construct PADs, which possess a larger coverage of the entire feature space, to create more potential overlaps with the target domain. The strategy is designed to significantly enhance the detection generalization ability.}
\label{fig:motivation}
\end{figure}

In this paper, we provide a new perspective to further analyze the generalization ability of adversarial detection methods. Clearly, the threat model of all different adversarial attacks should be identical to ensure fair analysis and comparison. Thus, the features extracted by the model from different attacks are of same dimensions within a shared feature space. Due to the variations in configurations such as attack objectives, parameters, and loss functions, the features of the examples generated from each attack form a distinct domain, named Adversarial Domain~(AD), as depicted in Fig.~\ref{fig:motivation}(a). 
Formally, Adversarial Domain~(AD) of a particular adversarial attack is defined as the cumulative representations of all the adversarial examples, which are generated by that attack. Existing detection methods typically select a random single attack to generate the training samples, i.e., they only select one AD as the source domain, as shown in Fig.~\ref{fig:motivation}(b). Apparently, there are few intersections between the source and unseen target domains, which usually lead to poor generalization performance. Additionally, the randomness inherent in the selection of the source domains will induce considerable fluctuations in generalization performance. 

A straightforward solution to the above issues, as shown in Fig.~\ref{fig:motivation}(c), is to randomly select multiple ADs as the source domains. This strategy tends to create a larger overlap with the target domain(s) and thereby improves the generalization performance of adversarial detection. Nonetheless, this strategy also induces uncertainty, and the selected similar ADs incur additional training costs without enough performance gains.

To further address the aforementioned problems, we propose a novel detection method, named Adversarial Example Detection via Principal Adversarial Domain Adaptation (AED-PADA). As shown in Fig.~\ref{fig:motivation}(d), by selecting Principal Adversarial Domains (PADs) as the source domains, which significantly enlarges the coverage of the entire adversarial feature space and creates larger overlap with the target domain, we can offer superior generalization performance. 

Specifically, AED-PADA contains two stages, i.e., Principal Adversarial Domains Identification (PADI) and Principal Adversarial Domain Adaptation (PADA). In the stage of PADI, since the discrepancies between the adversarial examples from various adversarial attacks are quite different compared to these of the ordinary classification tasks, we exploit adversarial supervised contrastive learning(Adv-SCL) to construct distinguishable ADs. Then, the selection of the most representative ADs must meet two key criteria. Firstly, there should be a clear distinction between candidate ADs to avoid redundancy caused by selecting similar ADs. Secondly, the combination of the candidate ADs should cover as much of the entire feature space as possible. To select the most representative ADs, we propose a Coverage of Entire Feature Space (CEFS) metric. With our CEFS metric, the formed PADs possess broad coverage of the entire feature space, and thus effectively improve the likelihood of capturing the location of the unseen target AD(s). 

In the stage of PADA, we pioneer to exploit the mechanism of Multi-source Unsupervised Domain Adaptation~(MUDA) to effectively utilize the rich knowledge acquired by PADs, to detect the unseen adversarial examples in the target domain. The framework of PADA is compatible with various existing MUDA methods. Since typical MUDAs only focus on extracting the semantic features from the spatial domain, we propose an adversarial feature enhancement module to extract features from both the spatial and frequency domains to construct a more comprehensive representation of adversarial examples.

Our major contributions can be summarized as follows.
\begin{itemize}
    \item We propose a novel adversarial example detection method, named Adversarial Example Detection via Principal Adversarial Domain Adaptation, to significantly improve the generalization performance of adversarial detection.
    \item We propose Principal Adversarial Domains Identification to identify the PADs, which possess a large coverage of the entire adversarial example feature space, with the help of the constructed AD clustering and proposed CEFS metric.
    \item We propose Principal Adversarial Domain Adaptation for detecting adversarial examples, by exploiting adversarial feature enhancement based Multi-source Unsupervised Domain Adaptation~(MUDA), which is compatible with various existing MUDA methods. To the best of our knowledge, this is the first work to exploit MUDA for adversarial example detection. 
\end{itemize}

\section{Related Work}

\subsection{Adversarial Example Detection}
The majority of existing adversarial example detection methods rely on statistical features~\cite{HendrycksG17,magnet,FarAwayDis1,FarAwayDis2,Hidden4,Erase-and-Restore}. They usually assume that the benign and adversarial examples originate from different distributions, and construct detectors based on the distinct statistical characteristics of these examples. Specifically, Grosse \emph{et al.}~\cite{STest1} utilize Maximum Mean Discrepancy for the adversarial example detection. Li \emph{et al.} employ Principal Component Analysis (PCA) to extract statistical feature, and construct a cascade classifier based on Support Vector Machines (SVMs)~\cite{PCAInconsistency}. Feinman \emph{et al.}~\cite{KDBU} carry out the detection based on Kernel Density (KD) and Bayesian-Uncertainty (BU) estimation. Ma \emph{et al.}~\cite{lid} exploit the concept of Local Intrinsic Dimensionality (LID) to calculate the distance between the distribution of inputs and their neighbors. Lee \emph{et al.}~\cite{md} utilize Guassian Discriminant Analysis (GDA) to model the difference between the benign and adversarial samples, and differentiate them based on Mahalanobis Distance (MD). Liu \emph{et al.}~\cite{steg} point out that steganalysis could be applied to adversarial example detection, and propose a steganalysis-based detection method (Steg). Tian \emph{et al.}~\cite{SID} reveal the inconsistency in the boundary fluctuations between the adversarial and benign examples, and construct Sensitivity Inconsistency Detector (SID) to identify the adversarial examples. Wang \emph{et al.}~\cite{txt_advdetection} embed hidden-layer feature maps of DNNs into word vectors, and detect adversarial examples via Sentiment Analysis~(SA). 

Existing adversarial detection methods exhibit poor generalization because their training typically depends on a single known attack which vastly differs from unseen test attacks. In this paper, we propose a novel adversarial detection method, which can substantially increase the coverage of the entire adversarial feature space and create larger overlap with the test adversarial attacks.

\subsection{Multi-source Unsupervised Domain Adaptation}

Transfer learning~\cite{transfer_learning_survey1,transfer_learning_survey2,transfer_learning_survey3} is a deep learning technique which leverages knowledge acquired from the source task(s) to improve learning efficiency and performance on a related but different target task. Unsupervised domain adaptation (UDA)~\cite{unsuperviesed_domain_adaptation1} is a type of popular method in transfer learning which aims to migrate knowledge learned from the labeled source domain(s) to the target domain, where only unlabeled target data are available for training. Single-source Unsupervised Domain Adaptation~(SUDA)~\cite{single-uda1,single-uda2,single-uda3} is widely explored in the previous research, which can transfer knowledge from one single source to one target domain. Compared to SUDA, Multi-source Unsupervised Domain Adaptation~(MUDA) acquires richer information while introduces a new challenge, i.e., how to effectively bridge the domain gaps between all source domains and the target domain.  

Various distribution alignment schemes have been proposed to achieve alignment between source and target domains. For example, Multiple Feature Spaces Adaptation Network~(MFSAN)~\cite{aaai-mda} leverages Maximum Mean Discrepancy~(MMD) to align the distributions of each pair of source and target domains in multiple specific feature spaces and aligns the outputs of classifiers by utilizing the domain-specific decision boundaries. Peng~\emph{et al.}~\cite{moment-mda} provide new theoretical insights specifically for moment matching to align the sources with each other and with the target. Owing to the development of generative adversarial networks, adversarial learning is widely used to find a domain-invariant feature space. It either focuses on approximating all combinations of pairwise domain discrepancies between each source and the target~\cite{cocktail,DARN} or uses a single domain discriminator~\cite{MIAN}. Other explicit measures of discrepancy, such as Wasserstein distance~\cite{mdmn,Wasserstein}, are also employed in MUDA to align the distribution of features. In addition to distribution alignment, the graph-matching metric~\cite{graph-muda1,graph-muda2} also considers the structural and geometric information, which achieves the alignment between the source and target domains by mapping both nodes and edges in a graph.

In this paper, we argue that the poor generalization performance of adversarial detection is due to the significant discrepancy between the source domains utilized for training and the target domain utilized for testing. Therefore, based on the MUDA approach, we propose a viable solution, Principal Adversarial Domain Adaptation, to reduce this great gap.

\section{Methodology}
\begin{figure*}[t]
\centering
\includegraphics[width=0.99\textwidth]{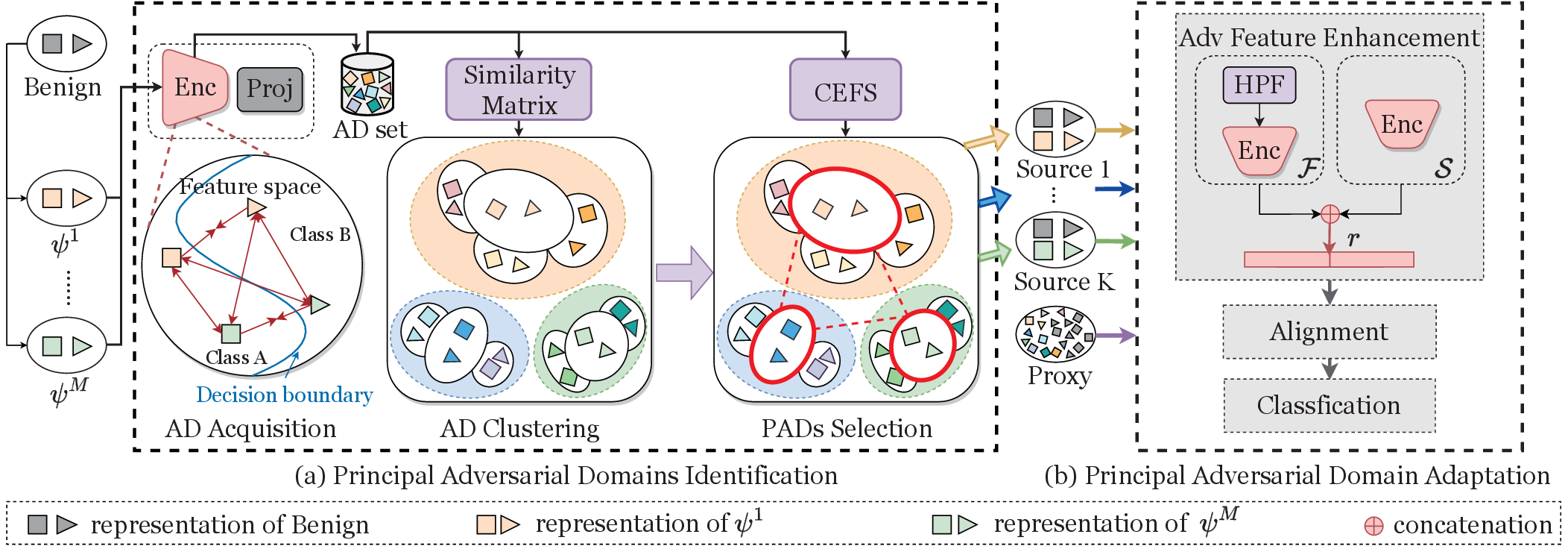} 
\caption{Our AED-PADA framework contains two stages: (a) Principal Adversarial Domains Identification, which consists of Adversarial Domain Acquisition, Adversarial Domain Clustering and Principal Adversarial Domains Selection, and (b) Principal Adversarial Domain Adaptation.}
\label{fig:framework}
\end{figure*}

To improve the generalization ability of adversarial example detection, we propose a novel adversarial example detection method, named Adversarial Example Detection via Principal Adversarial Domain Adaptation (AED-PADA). The entire framework of our AED-PADA is shown in Fig.~\ref{fig:framework}. AED-PADA contains two stages, Principal Adversarial Domains Identification (PADI) and Principal Adversarial Domain Adaptation (PADA). 
In the stage of PADI, we first incorporate adversarial supervised contrastive learning(Adv-SCL) to acquire distinguishable ADs. Then, we construct AD clustering to group ADs into different clusters. By proposing the Coverage of Entire Feature Space (CEFS) metric, we select the most representative ADs from each cluster to form PADs. In the stage of PADA, we propose an adversarial feature enhancement method based on the original MUDA method to effectively leverage PADs to detect the unseen adversarial attack methods. Note that Secs. \ref{sec:ad acquisition}, \ref{sec:ad clustering}, \ref{sec: pads confirming} and \ref{sec:pseudo padi} introduce PADI while Sec.~\ref{pada} presents PADA.

\subsection{Notations}
Suppose we have a labeled $C-$class classification dataset $\mathbb{D}=\{(x_i,y_i)\}_{i=1}^N$ with $N$ samples, where the label $y \in \{1,2,\ldots,C\}$. A classifier $f$ is trained on $\mathbb{D}$ to classify an input sample into one of the $C$ classes, $f(x) \rightarrow \mathbb{Z}_C$. Adversarial attack $\psi$ aims to fool $f$ into assigning incorrect labels via generating adversarial examples. We have a set of adversarial attack methods, $\Psi=\{\psi^m\}_{m=1}^M$, which consists of $M$ distinct adversarial attack methods. $\mathbb{D}^m=\{(x_i^m,y_i^m)\}_{i=1}^N$ is the adversarial dataset generated by the $m$-th adversarial attack method $\psi^m$, where $x_i^m$ denotes the $i$-th adversarial example generated by the $m$-th adversarial attack $\psi^m$, and $y_i^m$ presents its corresponding prediction label. We define $\varphi$ as, $\varphi(x_i^m)=\psi^m$, which is a  mapping from an adversarial example to its attack method. Consequently, we construct a set of adversarial examples $\mathcal{D}$ based on $\mathbb{D}$, which comprises $M$ types of adversarial examples, $\mathcal{D}=\{\mathbb{D}^m\}_{m=1}^M$.

\subsection{Adversarial Domain Acquisition} \label{sec:ad acquisition}

Typically, when we extract the features of the adversarial examples, which are generated from different untargeted attacks via common CNNs, the features tend to spread in an indistinguishable manner in the feature space. To acquire distinguishable representations of adversarial examples from different attacks, we exploit adversarial supervised contrastive learning~(Adv-SCL) to extract features. Then, we form the Adversarial Domains~(ADs), each of which is defined as the representations of all the adversarial examples generated from a particular adversarial attack.

Based on the supervised contrastive learning (SCL) method~\cite{KhoslaTWSTIMLK20}, Adv-SCL neglects the classification result of the  adversarial examples and focuses sorely on identifying their generation methods. Specifically, each adversarial example $x_i^m$ in $\mathcal{D}$ is characterized by the method used to generate it, i.e., the adversarial attack method $\psi^m$. The $\psi^m$ of the adversarial example $x_i^m$ serves as a key discriminant for determining whether different $x_i^m$ are positive or negative samples. Here, a pair of examples from the same attack are considered as positive, while those from different attacks are considered as negative. This learning strategy amplifies the dissimilarities across examples from various attacks and generates a more appropriate representation.

As shown in Fig.~\ref{fig:framework}(a), the input of AD acquisition is the adversarial example set $\mathcal{D}$. Adv-SCL consists of an Encoder Network ($Enc$) and a Projection Network ($Proj$). $Enc(\cdot)$ extracts a feature vector from the input adversarial example $x_i^m$, and $Proj(\cdot)$ further projects this representation vector to an auxiliary vector $z_i=Proj(Enc(x_i^m))$, which will be discarded after training.

\cite{transforming1,transform2} investigate that transformation strategies, such as cropping and rescaling, bit-depth reduction, JPEG compression, and randomization, have been used to defend against adversarial examples. Therefore, in order to prevent any potential ineffectiveness of adversarial samples caused by transformations, we only use normalization ($Norm$) as the data augmentation operation in the stage of AD Acquisition. 

We randomly sample $n$  example-label pairs in $\mathcal{D}$, $\{(x_k^{m_k},y_k^{m_k})\}_{k=1}^n$, where $m_k$ is the adversarial attack of the $k$-th adversarial example. The corresponding batch employed for training consists of $2n$ pairs, $\{(\tilde{x}_l,\tilde{y}_l)\}_{l=1}^{2n}$, where $\tilde{x}_{2k}$ and $\tilde{x}_{2k-1}$ are two views of $x_k^{m_k}$, and they are from the same adversarial attack method $\psi^{m_k}$. The loss function of Adv-SCL is defined as,
\begin{equation}
    \mathcal{L}^{sup} = \sum_{i \in I} \mathcal{L}_i^{sup},
    \label{eq:sup_contrastive}
\end{equation}

\begin{equation}
    \mathcal{L}_i^{sup}=\frac{-1}{|P(i)|} \sum_{p \in P(i)} \log \frac{\exp (z_i \cdot z_p / \tau)}{\sum_{a \in A(i)} \exp (z_i \cdot z_a / \tau)}.
\end{equation}
Here, $i \in I = \{1,\ldots,2n\}$ denotes the index of the augmented samples. $A(i)$ is a subset of $I$ which includes all indexes except $i$. $P(i) =\{p \in A(i) : \varphi(\tilde{x}_p) = \varphi(\tilde{x}_i)\}$ represents the indices of positive samples in the batch except $i$, and $|P(i)|$ stands for its cardinality. $\cdot$ denotes the dot product. $\tau$ is the temperature parameter which scales the similarity values.

\subsection{Adversarial Domain Clustering} \label{sec:ad clustering}

After the acquisition of ADs, we obtain $\mathbb{H}^m=Enc(\mathbb{D}^m)$ as AD of the corresponding adversarial attack $\psi^m$. For $\mathcal{D}$, we construct a set of ADs, $\mathcal{H}=\{\mathbb{H}^m\}_{m=1}^M$. 
Since different ADs tend to distribute differently in the feature space and many of them possess various portions of overlaps, it is quite difficult to directly select the most representative ADs from scratch. Then, it is vital to explore the similarities among different ADs.

To address this issue, we intend to perform the selection via two steps, i.e., clustering to assess the similarities of ADs and selecting the most representative ADs from each cluster. Then, we construct a viable solution named Adversarial Domain Clustering, as shown in Fig. \ref{fig:framework}(a). This strategy groups ADs into different clusters, by ensuring that the similarity among ADs within the same cluster is maximized, while the similarity among ADs across different clusters is minimized. AD clustering avoids the redundancy and additional costs by preventing the repeated selection of similar attack methods, and it facilitates the selection of the most representative ADs in the subsequent steps.

Since the samples to be clustered here are collections of features, rather than individual data points, traditional clustering methods such as K-Means~\cite{kmeans} cannot be directly applied. Since spectral clustering \cite{NgJW01} only requires the similarity matrix among samples, it is utilized to construct the clustering step in our AD clustering.

For the estimation of the similarity matrix $W \in R^{M \times M}$ in spectral clustering, which represents the similarities between different ADs, we propose Adversarial Domain Similarity Measurement (ADSM) based on Jensen-Shannon divergence (JSD)~\cite{ErvenH14}, which quantifies the similarity between two probability distributions. To compute the similarities, we transform each $\mathbb{H}^i$ in $\mathcal{H}$ to $\overline{\mathbb{H}^i}$ by converting $\mathbb{H}^i$ into probabilities and performing normalizations. Since a smaller value of JSD between two ADs implies a smaller discrepancy in their probability distributions, $\text{ADSM}(\mathbb{H}^i,\mathbb{H}^j)$ can be computed via 
\begin{equation}
    \text{ADSM}(\mathbb{H}^i,\mathbb{H}^j)=\frac{1}{\text{JSD}(\overline{\mathbb{H}^i},\overline{\mathbb{H}^j})}.\label{eq:adsm}
\end{equation}

By letting the element $W_{i,j}$ at the $i$-th row and $j$-th column refer to the similarity between $\mathbb{H}^i$ and $\mathbb{H}^j$, $W_{i,j}$ can be calculated by
\begin{equation}
    W_{i,j} = \left\{\begin{array}{cc}
      \text{ADSM}(\mathbb{H}^i,\mathbb{H}^j)   & i \neq j \\ 
       0  & i = j
    \end{array}. 
    \right. \label{eq:wij}
\end{equation}

Since the spectral clustering cannot automatically determine the optimal number of clusters, the Calinski-Harabasz score (CH score)~\cite{mann2023proposed}, which requires no knowledge of the cluster shape, is utilized to evaluate the clustering performance and estimate the optimal number of clusters. Note that it measures both the within-cluster and between-cluster distances, thereby offering a more comprehensive view of the clustering performance. CH score is calculated by
\begin{equation}
    \text{CH}(K) = \frac{Tr(BC_K) / (K-1)}{Tr(WC_K) / (N-K)}, \label{eq:ch_score}
\end{equation}
where $K$ and $N$ is the number of clusters and data respectively, $BC_K$ denotes between-cluster covariance matrix, $WC_K$ denotes within-cluster covariance matrix, and $Tr(\cdot)$ denotes the trace of the matrix. 

Higher value of $\text{CH}(K)$ indicates better clustering. We posit that the inherent structure of the entire feature space composed of different ADs is highly complex. Given that the CH score often awards the highest evaluation when cluster numbers $K=2$, we choose to commence our consideration from the scenario where $K=3$ in this paper. With the help of the CH score, our AD clustering can automatically group the ADs into optimal number of clusters.

\subsection{Principal Adversarial Domains Selection} \label{sec: pads confirming}

After similar ADs are clustered, then the selection step can be performed. To form the most effective Principal Adversarial Domains (PADs), it is vital to select appropriate ADs from different clusters. Thus, we propose the Coverage of Entire Feature Space metric (CEFS) to guide the PADs selection process. 

CEFS is a ratio-based metric which contains two aspects, Intra-Domain Dispersion (IDD) and Discrepancy between Adversarial Domains (DAD). IDD represents the dispersion among features within each AD, where a higher value indicates a larger coverage of the feature space. For any AD, $\mathbb{H}^i = \{{h_1^i},{h_2^i},\ldots,{h_v^i}\} \in R^{v \times d}$, where $v$ and $d$ denote the number and dimension of features in $\mathbb{H}^i$, respectively, IDD can be computed by 
\begin{equation}
    \text{IDD}(x) = \frac{1}{v} \sum_{i=1}^{v} \left\| \frac{x_i}{\|x_i\|_2} - \frac{1}{v} \sum_{j=1}^{v} \frac{x_j}{\|x_j\|_2} \right\|_2. \label{eq:IDD}
\end{equation}

DAD represents the discrepancies between the two selected ADs, this discrepancy can be quantified using a distance metric $Dist(\cdot,\cdot)$, such as Kullback-Leibler divergence~\cite{kl-divergence} or Maximum Mean Discrepancy (MMD)~\cite{mmd}. A lower DAD value indicates a greater similarity between the two selected ADs. DAD can be calculated as,
\begin{equation}
    \text{DAD}(\mathbb{H}^i,\mathbb{H}^j)=Dist(\overline{\mathbb{H}^i},\overline{\mathbb{H}^j}). \label{eq:DAD}
\end{equation}

Then, CEFS can be obtained via 
\begin{equation}
    \text{CEFS} = \frac{\sum_{i=1}^K \text{IDD}(\mathbb{H}^i)}{\text{DAD}((\mathbb{H}^1 \parallel \cdots \parallel\mathbb{H}^K),(\mathbb{H}^1 \parallel \cdots \parallel\mathbb{H}^M))}, \label{eq:cefs}
\end{equation}
where $K$ and $M$ is the number of ADs in PADs and the number of ADs in the AD set $\mathcal{H}$ respectively, and $\parallel$ denotes a concatenation operation. 

CEFS is a ratio-based metric to quantify the coverage of selected ADs within the entire feature space (EFS). In Eq.~\eqref{eq:cefs}, the numerator represents Intra-Domain Dispersion (IDD), indicates feature dispersion within each AD. The denominator measures the discrepancy between the selected ADs and EFS. As CEFS increases, the numerator grows, indicating a larger feature space for each AD, and the denominator decreases, suggesting a greater similarity between the selected ADs and the EFS. Consequently, the selected ADs have a larger coverage of the EFS, increasing the likelihood of capturing unseen ADs. We utilize CEFS to select PADs, which can give a larger coverage of the feature space with the same number of ADs. PADs actually enhance the probability of capturing the location of unseen ADs, thereby improving the generalization performance.

\subsection{Training process of Principal Adversarial Domains Identification} \label{sec:pseudo padi}
The process of Principal Adversarial Domains Identification (PADI) consists of Adversarial Domain Acquisition (AD Acquisition), Adversarial Domain Clustering (AD Clustering) and Principal Adversarial Domains Selection (PADs Selection). The training process of Principal Adversarial Domains Identification is described in Algorithm~\ref{alg:PADI}.

\begin{algorithm*}[!t]
    \caption{Training process of Principal Adversarial Domains Identification}
    \begin{algorithmic}[1]
    \Require $\mathcal{D}_{acq},\mathcal{D}_{clu}$.
    \Ensure Principal Adversarial Domains~(PADs).
    \Statex $\triangleright$ Adversarial Domain Acquisition
    \For{$x_i^{m_i}$ $\mathbf{in}$ $\mathcal{D}_{acq}$}
        \State Augment $x_i^{m_i}$ to be $(\tilde{x}_{2i-1},\tilde{x}_{2i})$ with augmentation $Norm$
        \State $z_i = Enc(Proj(\tilde{x}_i))$
        \State Get $\mathbf{Enc}$, $\mathbf{Proj}$ via minimizing $\mathcal{L}$ in Eq.~\eqref{eq:sup_contrastive}
    \EndFor
    
    \Statex $\triangleright$ Adversarial Domain Clustering 
    \State Generate the AD pool $\mathcal{H}_{clu} $ based on $\mathcal{D}_{clu}$
\State Generate the similarity matrix $W$ by Eq.~\eqref{eq:adsm}
    \For { $k$ $\mathbf{in}$ [3,$M$]}
        \State $r_k = SC(W,k)$ \Comment{ $r_k$ is the clustering result via Spectral clustering.}
        \State Get $\text{CH}(k)$ by Eq.~\eqref{eq:ch_score}
    \EndFor
    \State $K=\operatorname*{arg\,max}_k (\text{CH}(k))$ \Comment{$K$ is the best number of clusters.}
    \State $r_K=SC(W,K)=\{clu_i\}_{i=1}^K$ \Comment{$r_K$ is the best clustering result which contains $K$ clusters.} 
    \Statex $\triangleright$ Principal Adversarial Domains Selection
    \State $C=\Pi_{i=1}^Kn_i$ \Comment{$n_i$ is the number of adversarial attacks in $clu_i$, and $\Sigma_{i=1}^Kn_i=M.$}
    \State Generate combination list $L=\{l_i\}_{i=1}^C$ \Comment{$L$ is the list of all combinations in $r_K$, each $L_i$ has $K$ types of attacks.}
    \For{$l_i$ $\mathbf{in}$ $L$
    }
        \State Generate $\mathcal{H}_i=\{\mathbb{H}_i^j\}_{j=1}^K$ based on $l_i$\Comment{$\mathcal{H}_i$ is the list of ADs based on $l_i$}
        \State Get CEFS($\mathcal{H}_i$) by Eq.~\eqref{eq:cefs}
    \EndFor
    \State $\text{PADs}=\operatorname*{arg\,max}_{l_i}(\text{CEFS}(\mathcal{H}_i))$

    \end{algorithmic}
    \label{alg:PADI}
\end{algorithm*}

For the original dataset $\mathbb{D}$, we divide it into two non-overlapping datasets $\mathbb{D}_{acq}$ and $\mathbb{D}_{clu}$, which are used for AD Acquisition and AD Clustering, respectively. The adversarial examples sets for AD Acquisition are denoted as 
$\mathcal{D}_{acq}$. $\mathcal{D}_{acq}=\{\mathbb{D}_{acq}^m\}_{m=1}^M, \mathbb{D}_{acq}^m =\{(x_i^m,y_i^m)\}_{i=1}^{N_{m}}$, where $M$ is the number of adversarial attacks, $\mathbb{D}_{acq}^m$ is the adversarial examples set generated by the $m$-th adversarial attack, and $N_m$ is the cardinality of $\mathbb{D}_{acq}^m$.

Likewise, The adversarial examples sets for AD Clustering are denoted as 
$\mathcal{D}_{clu}$. $\mathcal{D}_{clu}=\{\mathbb{D}_{clu}^q\}_{q=1}^M$, $\mathbb{D}_{clu}^q =\{(x_i^q,y_i^q)\}_{i=1}^{N_{q}}$, where $M$ is the number of adversarial attacks, $\mathbb{D}_{clu}^q$ is the adversarial examples set generated by the $q$-th adversarial attack, and $N_q$ is the cardinality of $\mathbb{D}_{clu}^q$. The rationale behind this strategy is to prevent the model after contrastive learning from overfitting to the data on AD acquisition, which leads to subpar performance in AD Clustering and PAD Confirming.

\subsection{Principal Adversarial Domain Adaptation} \label{pada}

To transfer the learned knowledge from PADs to the target domain, i.e., the adversarial examples generated from unseen methods, we propose Principal Adversarial Domain Adaptation~(PADA) to detect adversarial examples, as depicted in Fig.~\ref{fig:framework}(b).

Ideally, the inputs of PADA comprise source data and unseen target data. The source data, denoted as, $Src=\{Src_i\}_{i=1}^K$, consists of an equal number of benign examples and adversarial examples, where the adversarial examples contain $K$ types of adversarial examples determined by PADs. Since the unseen target data is unavailable, we can only use the training data as a proxy. The proxy data, denoted as $PD$, also contains an equal number of benign examples and adversarial examples, drawn from the training benign set $\mathbb{D}$ and the training adversarial set $\mathcal{D}$, respectively. $\mathcal{D}$ contains $M$ types of adversarial examples. During the PADI stage, we select the most representative ADs, i.e., $K$ types of ADs from these $M$ types to form the PADs. Consequently, employing $\mathcal{D}$ as a proxy for unseen target data serves two key purposes. Firstly, it prevents overlap between the source data used for training and the unseen target data used for testing. Secondly, the PADA process compels the transfer of specific knowledge from PADs to the more extensive set $\mathcal{D}$. This transfer aims to boost the generalization capabilities of networks of PADA, with the goal of enhancing their performance when testing the unseen target data.

PADA consists of three sequential components, feature extraction, feature alignment and classification. The framework of PADA is compatible with various widely used Multi-source Unsupervised Domain Adaptation~(MUDA) methods~\cite{moment-mda,DARN,mdmn,aaai-mda}. The experimental results indicate that our PADA possesses excellent generalization capabilities based on various existing MUDA methods. Due to the simplicity and effectiveness of MFSAN~\cite{aaai-mda}, along with its superior detection performance compared to other MUDA methods, we select MFSAN as the basic MUDA method for our PADA.

\begin{figure}[!t]
    \centering
    \includegraphics[width=0.65\columnwidth]{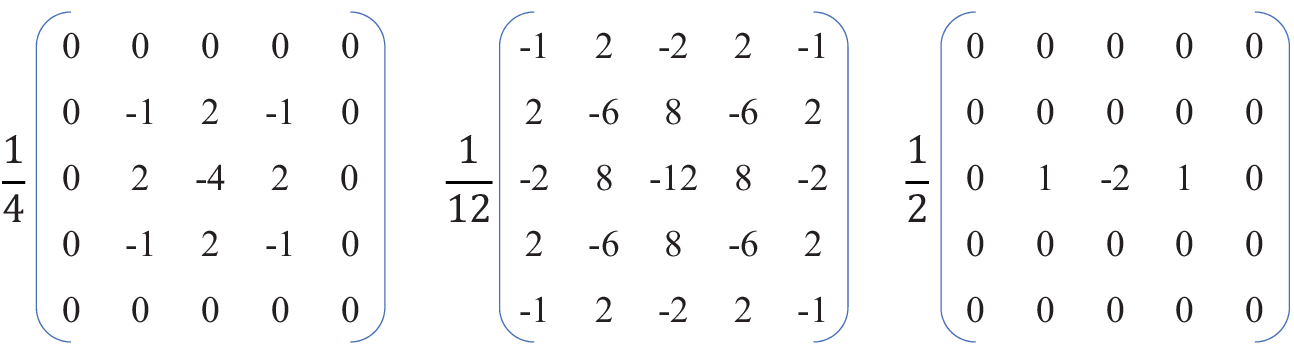} 
    \caption{The three kernels of PEF employed to capture the perturbation signals hidden in the adversarial inputs.}
    \label{fig:SRM}
\end{figure}

In the feature extraction component, unfortunately, existing MUDA methods including MFSAN, only extract spatial features. \cite{GeirhosRMBWB19,WangWHX20,FanLCZG21} indicate that the high-frequency component of an image plays a crucial role in the prediction of deep neural network. Adversarial perturbations are more likely to be concealed in the high-frequency information of images. To capture more comprehensive features of adversarial examples, we propose an adversarial feature enhancement~(AFE) module as the feature extraction component. AFE contains both the spatial feature extraction~$\mathcal{S}$ and frequency feature extraction~$\mathcal{F}$ branches. For frequency feature extraction, we design perturbations extraction filters (PEF), which is a plug-and-play operation based on Spatial Rich Model~(SRM)~\cite{SRM} to capture subtle perturbation signals hidden in the adversarial examples.

SRM typically uses 30 basic kernels to capture textures and discontinuities of images, and has been employed in image forensics to detect subtle and irregular manipulation or hidden information. Subsequent research~\cite{SRM_detection} indicates that in image manipulation detection, employing only three kernels can achieve considerable detection performance, and more many basic kernels does not further enhance performance. Essentially, these kernels are high-pass filters, which enhance the high-frequency signals and remove the low-frequency components of the inputs. Adversarial perturbations are typically hidden within the high-frequency information of an image. Therefore, our PEF uses the same three kernels to extract subtle perturbation signals. As shown in Fig.~\ref{fig:SRM}, PEF consists of three filters which are implemented by convolution kernels with fixed parameters. We set the kernel size of PEF to be $5\times5\times3$, and the output channel size of PEF is 3. Both $\mathcal{S}$ and $\mathcal{F}$ employ $Enc$ from Sec.~\ref{sec:ad acquisition}, aligned with the threat models, specifically ResNet-18 or VGG-16. Then, the enhanced adversarial features, denoted as $r(x)=[\mathcal{S}(x),\mathcal{F}(x)]$, are fed into the next module. 

In the alignment component, we assign the specific network $Q_i$ to map each source domain $Src_i$ and proxy domain $PD$ to the different feature space, and utilizes MMD as the distance metric to align them. $L_d$ is used to align the features between source domain and proxy domain, 
\begin{equation}
    L_d = \frac{1}{K}\sum_{i=1}^K\text{MMD}(Q_i(r(Src_i)),Q_i(r(PD))),
\end{equation}
where $r$ is the enhanced adversarial features after AFE, $K$ is the number of source domains. each source domain is associated with the corresponding network $Q_i$.

In the classification component, we utilize the cross-entropy loss $L_{cls}$ to ensure the correct classification. Given that different domain-specific classifiers~$C_i$ are trained on their respective source domains, resulting in significant discrepancies among their predictions for the same proxy domain. To address this, $L_{disc}$ is used to minimize the differences among the predictions of various classifiers.

\begin{align}
    L_{disc} = &\frac{2}{K\times(K-1)}\sum_{j=1}^{K-1}\sum_{i=j+1}^{K}\mathbb{E}_{x\sim PD} 
    |C_i(Q_i(r(x)))-C_j(Q_j(r(x)))|.
\end{align}

Overall, the total loss is formulated as follow, $\lambda$ and $\gamma$ are hyperparameters used to adjust the weights of $L_d$ and $L_{disc}$, respectively.

\begin{equation}
    L_{total} = L_{cls} + \lambda L_d + \gamma L_{disc}.
\end{equation}

\section{Experiments}

\subsection{Experimental Setups}

\subsubsection{DNN backbones}
We evaluate the performance of the proposed method, by employing two widely used DNN architectures, i.e., ResNet-18 ~\cite{HeZRS16} and VGG-16~\cite{SimonyanZ14a}, according to~\cite{SID}. 

\subsubsection{Datasets}

We evaluate the performance of the proposed method on three popular datasets, CIFAR-10~\cite{cifar10}, SVHN~\cite{SVHN} and ImageNet~\cite{imagenet}. All the images in ImageNet are resized to $224 \times 224 \times 3$ via pre-processing.

As shown in Table~\ref{table:dataset_splitting}, each of the three datasets is divided into three category-balanced and non-overlapping subsets: \textit{Train-acq-src}, \textit{Train-clu-pro} and \textit{Test}. The training data for AD Acquisition in the PADI stage and the source data in the PADA stage are selected from \textit{Train-acq-src}. Similarly, the training data for AD Clustering in the PADI stage and the proxy data in the PADA stage are selected from \textit{Train-clu-pro}.

For CIFAR-10, we divide the official CIFAR-10 training set into two halves, each of which contains 25,000 images, to form \textit{Train-acq-src} and \textit{Train-clu-pro}. Besides, we form \textit{Test} with all the 10,000 images in the official CIFAR-10 testing set. For SVHN, we set the number of \textit{Train-acq-src} and \textit{Train-clu-pro} to 20,000 instead of 25,000, due to the presence of category imbalance in the official SVHN training dataset, to construct a category-balanced training dataset. \textit{Test} of SVHN also consists of 10,000 images, which are randomly selected from the official SVHN testing set. For ImageNet, We divide the official ImageNet (ILSVRC2012) validation set, which consists of 50,000 images, into two parts, 40,000 images for training and 10,000 images for \textit{Test}. The 40,000 training images are then equally divided into two subsets, \textit{Train-acq-src} and \textit{Train-clu-pro}, each of which contains 20,000 images.

Ten types of training adversarial examples are created based on the benign examples from the \textit{Train-acq-src} and \textit{Train-clu-pro} subsets. Subsequently, another seven attacks are applied to generate the testing adversarial examples on the \textit{Test} set. These specific adversarial attack methods are detailed in Sec.~\ref{sec:baseline attack}. Note that the attack methods for training and testing are entirely different, ensuring that the training data and testing data are mutually exclusive.

\subsubsection{Data splitting strategy for the training stage}

Table~\ref{table:trainging_stage} presents the data splitting strategy for the training stage of our AED-PADA. AED-PADA contains two stages during training. In the PADI stage, to improve the efficiency, we randomly select 10,000 images from each type of adversarial attacks and their corresponding benign images from the \textit{Train-acq-src} for supervised contrastive learning. To alleviate the overfitting problem, we select 10,000 images per attack from \textit{Train-clu-pro} for AD Clustering and PAD Confirming.

In the PADA stage, the inputs of PADA are the images from the source domains and the unseen target domains. Each source domain contains 10,000 samples, with an equal split of 5,000 benign examples and 5,000 adversarial examples, which are all randomly selected from \textit{Train-acq-src}. Since the data from unseen domains is unavailable, we can only utilize the training examples as a proxy, which are generated from all the 10 types of training attacks. Specifically, the proxy domain contains 10,000 samples, i.e., 5,000 benign examples and 5,000 adversarial examples, which are all randomly selected from the \textit{Train-clu-pro}. Note that the adversarial examples in the proxy domain are obtained via 10 different training attacks, with each attack providing 500 examples. This specific arrangement has two benefits. Firstly, it ensures zero overlap between the source and proxy domains during training, to avoid data leakage. Secondly, it enhances the diversity of the training data, to further benefit the generalization of the detection model.

\begin{table}[!t]
    \centering
    \caption{Datasets for training and testing.}
    
    \resizebox{0.70\columnwidth}{!}{\begin{tabular}{c|cc|cc|cc}
        \hline
        \multirow{2}{*}{Dataset} &\multicolumn{2}{c|}{Train-acq-src}  &\multicolumn{2}{c|}{Train-clu-pro} &\multicolumn{2}{c}{Test} \\
        & benign &adv &benign &adv &benign &adv \\
        \hline
        \rule{0pt}{9pt}
        CIFAR-10 &$25,000$ &$10\times25,000$ &$25,000$ &$10\times25,000$ &$10,000$ &$7\times10,000$\\
        SVHN &$20,000$ &$10\times20,000$ &$20,000$ &$10\times20,000$ &$10,000$ &$7\times10,000$ \\
        ImageNet &$20,000$ &$10\times20,000$ &$20,000$ &$10\times20,000$ &$10,000$ &$7\times10,000$ \\
        \hline
    \end{tabular}}
    \label{table:dataset_splitting}
\end{table}

\begin{table}[!t]
    \centering
    \caption{Data splitting for the training stage of our AED-PADA.}
    
    \resizebox{0.70\columnwidth}{!}{\begin{tabular}{cccc}
        \hline
        \multicolumn{2}{c}{Training stage} & benign examples & adversarial examples \\
        \hline
        \multirow{2}{*}{PADI} &AD Acquisition &- &$10,000$/attack \\
        & AD Clustering &- &$10,000$/attack \\
        \hline
        \multirow{2}{*}{PADA} &source domain &$5,000$/source &$5,000$/source \\
        & proxy domain &$5,000$ &$5,000$ \\
        \hline

    \end{tabular}}
    \label{table:trainging_stage}
\end{table}

\subsubsection{Baseline adversarial attack methods} \label{sec:baseline attack}
To evaluate the generalization capabilities of the detection methods, it is important to consider a diverse set of attack methods, including both the earlier and recent techniques. Here, 10 earlier attack methods are utilized to generate adversarial examples for training, including FGSM~\cite{FGSM}, BIM~\cite{BIM}, C\&W~\cite{CW}, DeepFool~\cite{DeepFool}, PGD~\cite{pgd}, MI-FGSM~\cite{MIM}, DIM\cite{DIM}, ILA~\cite{ILA}, YA-ILA~\cite{ILA++} and SI-NI-FGSM~\cite{si-ni-fgsm}. 7 SOTA attack methods are employed to generate adversarial examples for testing, including APGD~\cite{APGD}, ILA-DA~\cite{ILA-DA}, Jitter~\cite{jitter}, SSA~\cite{SSA}, TI-FGSM~\cite{TIM}, VMI-FGSM~\cite{vmifgsm} and VNI-FGSM~\cite{vmifgsm}. The adversarial attacks for training and testing are entirely distinct, so the adversarial detection results on the unseen testing attacks indicate the generalization performance of our proposed detection method.

\subsubsection{Baseline adversarial detection methods}
We compare the generalizability of our AED-PADA with five state-of-the-art adversarial detection methods, LID~\cite{lid}, MD~\cite{md}, Steg~\cite{steg}, SID~\cite{SID} and SA~\cite{txt_advdetection}. These detection methods differ from ours as they employ only a single attack method for training. To ensure a fair comparison with our AED-PADA, it is necessary to consider the scenario training by multiple attacks. Consequently, we consider the following two configurations for the SOTA detection methods: (1) For single attack training, we utilize all data from \textit{Train-acq-src} to ensure sufficient data volume for effective training. (2) For multiple attacks training, the methods are trained by the adversarial examples in PADs, which are consistent with our AED-PADA. Each source domain contains 10,000 samples from \textit{Train-acq-src}, evenly divided into 5,000 benign and 5,000 adversarial examples.

\begin{table*}[!t]
    \centering
    \caption{Comparison of the generalization performances on CIFAR-10 between the state-of-the-art adversarial detection methods and our AED-PADA. Averaged Accuracy is the average result across seven unseen state-of-the-art testing adversarial attacks. The bolded and the underlined values represent the best and the second best results for each column, respectively.}

    \resizebox{.85\textwidth}{!}{\begin{tabular}{cccccccccc}
        \hline
        \rule{0pt}{10pt}
        \multirow{2}{*}{\makecell{Dataset \\ (Backbone)}}  & \multirow{2}{*}{Detector} & \multicolumn{7}{c}{Accuracy on Unseen SOTA Adversarial Attacks (\%)} & \multirow{2}{*}{\makecell{Averaged \\ Accuracy~(\%)}} \\
        \cline{3-9}
        \rule{0pt}{9pt}
         & &   APGD & \text{ILA-DA} & Jitter & SSA & TI-FGSM & VMI-FGSM & VNI-FGSM   \\
        \hline
        \rule{0pt}{9pt}
        \multirow{9}{*}{\makecell{CIFAR-10 \\ \\ (ResNet-18)}}  & LID~\cite{lid} & \underline{90.044} 	&94.121 & 78.240 & 59.839 & 90.863 & 91.056 & 86.496 & 84.380\\
        &  LID-PADs &78.903 & 86.068 & 70.045 & 57.330 & 82.278 & 82.333 	& 79.978 & 76.705 \\
        &  MD~\cite{md} & 65.976 & 96.657 & 61.764 & 54.225 & 69.282 & 69.653 & 69.726 & 69.612 \\
        &  MD-PADs & 63.371 & \textbf{97.970} & 62.058 & 51.501 & 68.100 & 68.814 & 69.351 & 68.738 \\
        &  Steg~\cite{steg} & 83.521 & 94.692 & 86.759 & 58.064 & 90.496 & 90.814 & 92.057 & 85.200 \\
        &  Steg-PADs & 84.085 & 94.155 & 88.885 & 62.305 & 90.715 & 90.860 & 91.690 & 86.099 \\
        &  SID~\cite{SID} & 86.113 & 60.999 & 74.629 & 53.732 & 87.323 & 87.330 & 82.100 & 76.032 \\
        &  SID-PADs & 86.520 & 61.820 & 74.860 & 54.010 & 88.285 & 88.220 & 82.935 & 76.664 \\
        &  SA~\cite{txt_advdetection} & 84.907 & 90.731 & 87.365 & 81.323 & 89.275 & 90.013 & 91.967 & 87.940 \\
        & SA-PADs & 89.125	& 93.485 & \underline{91.925}	& \textbf{88.885}	& \underline{94.760}	& \underline{94.875}	& \underline{95.595}	& \underline{92.664} \\
        & AED-PADA & \textbf{90.545} 	& \underline{97.855} 	& \textbf{97.065} 	& \underline{84.255} 	& \textbf{97.700} 	& \textbf{97.765} 	& \textbf{97.675} 	& \textbf{94.694} \\[1.3pt]
        \hline
        \rule{0pt}{9pt}
        \multirow{9}{*}{\makecell{CIFAR-10 \\ \\ (VGG-16)}}  & LID~\cite{lid} & \textbf{85.024} & \underline{93.357} & 75.092 & 58.639 & 88.201 & 88.173 & 81.540 & 81.432 \\
        &  LID-PADs & 70.493 & 89.403 & 65.190 & 58.588 & 75.553 & 75.408 & 73.500 & 72.590 \\
        &  MD~\cite{md} & 51.322 & 72.013 & 50.904 & 52.012 & 51.809 & 51.799 & 50.963 & 54.403 \\
        &  MD-PADs & 50.033 	& 81.779 & 50.028 & 51.349 	& 50.005 	& 50.856 	& 51.349 	& 55.057 \\
        &  Steg~\cite{steg} & 79.280 & 92.756 	& \underline{88.466} 	& 57.470 	& 88.854 	& \underline{88.891} 	& \underline{89.390} 	& 83.586 \\
        &  Steg-PADs & 79.900 & 92.265 	& 88.705 	& 59.360 	& \underline{89.015} 	& 88.800 	& 89.235 	& \underline{83.897} \\
        &  SID~\cite{SID} & 83.235 & 66.047 & 73.260 & 53.373 & 84.743 & 84.582 & 76.210 & 74.493 \\
        &  SID-PADs & 83.105 & 66.240 & 73.230 	& 53.375 	& 84.605 	& 84.385 	& 76.145 	& 74.441 \\
        & SA~\cite{txt_advdetection} & 80.198 	& 85.036 	& 81.003 	& \underline{63.984} 	& 85.356 	& 86.293 	& 87.627 	& 81.357 \\ 
        & SA-PADs & 79.720	& 83.825	& 77.985	& 63.335	& 83.765	& 84.475	& 85.755	& 79.873 \\
        & AED-PADA & \underline{84.370} 	& \textbf{94.365} 	& \textbf{93.260} 	& \textbf{65.765} 	& \textbf{93.135} 	& \textbf{93.410} 	& \textbf{93.545} 	& \textbf{88.264}  \\[1.3pt]
        \hline    
    \end{tabular}}

    \label{table:sota}
\end{table*}

\begin{table*}[!t]
    \centering
    \caption{Comparison of the generalization performances on SVHN between the state-of-the-art adversarial detection methods and our AED-PADA. Averaged Accuracy is the average result across seven unseen state-of-the-art testing adversarial attacks. The bolded and the underlined values represent the best and the second best results for each column, respectively.}

    \resizebox{.85\textwidth}{!}{\begin{tabular}{cccccccccc}
        \hline
        \rule{0pt}{10pt}
        \multirow{2}{*}{\makecell{Dataset \\ (Backbone)}}  & \multirow{2}{*}{Detector} & \multicolumn{7}{c}{Accuracy on Unseen SOTA Adversarial Attacks (\%)} & \multirow{2}{*}{\makecell{Averaged \\ Accuracy~(\%)}} \\
        \cline{3-9}
        \rule{0pt}{9pt}
         & &   APGD & \text{ILA-DA} & Jitter & SSA & TI-FGSM & VMI-FGSM & VNI-FGSM   \\
        \hline
        \multirow{9}{*}{\makecell{SVHN \\ \\ (ResNet-18)}}  & LID~\cite{lid} & 66.431 & 87.940 & 63.697 & 60.324 & 71.403 & 71.384 & 64.952 & 69.447 \\
        &  LID-PADs & 61.849	& 86.846	& 60.551	& 58.515	& 69.581	& 69.639	& 64.182	& 67.309  \\ 
        &  MD~\cite{md} & 61.167 & 66.439 & 59.906 	& 58.270 	& 67.625 	& 67.609 	& 60.870 	& 63.127  \\
        &  MD-PADs & 60.768	& 50.190	& 59.637	& 57.967	& 67.257 	& 67.337 	& 60.688 	& 60.549  \\
        &  Steg~\cite{steg} & 71.591 	& 51.509 	& 94.268 	& 59.865 	& 96.943 	& 96.874 & 95.954 & 81.000 \\
        &  Steg-PADs & \underline{73.460} 	& 50.860 	& \underline{97.115} 	& \underline{74.360} 	& \underline{98.005} 	& \underline{98.040} 	& \underline{97.575} 	& \underline{84.202} \\ 
        &  SID~\cite{SID} & 68.859 	& \underline{92.207} 	& 63.763 	& 56.571 	& 74.074 	& 73.995 	& 64.150 	& 70.517  \\
        &  SID-PADs  & 69.670 	& 93.420 	& 64.155 	& 56.800 	& 75.285 	& 75.120 	& 64.780 	& 71.319  \\
        & SA~\cite{txt_advdetection} & 68.092 & 73.066 & 65.288 & 69.276 & 82.758 & 82.823 & 82.467 & 74.824 \\ 
        & SA-PADs & 69.490 & 73.100 & 63.610 & 66.550 & 78.155 & 78.220 & 77.785 & 72.416 \\

        & AED-PADA & \textbf{74.455} 	& \textbf{99.730} 	& \textbf{99.730} 	& \textbf{87.305} 	& \textbf{99.730} 	& \textbf{99.730} 	& \textbf{99.730} 	& \textbf{94.344}  \\[1.0pt]
        \hline
        \rule{0pt}{9pt}
        \multirow{9}{*}{\makecell{SVHN \\ \\ (VGG-16)}} & LID~\cite{lid} & 65.645 	& 96.605 	& 62.731 	& 64.252 	& 74.205 	& 74.263 	& 70.296 	& 72.571  \\
        &  LID-PADs & 69.710 	& 97.154 	& 63.626 	& 63.955 	& 76.258 	& 76.306 	& 70.119 	& 73.875 \\
        &  MD~\cite{md} & 51.511 	& 58.333 	& 51.233 	& 51.859 	& 51.870 	& 52.268 	& 51.465 	& 52.648   \\
        &  MD-PADs & 51.920 	& 56.484 	& 51.578 	& 52.745 	& 51.103 	& 50.285 	& 51.748 	& 52.266  \\
        &  Steg~\cite{steg} & 70.586 	& 98.661 	& 93.882 	& 58.496 	& 95.774 	& 95.575 	& 94.549 	& 86.789  \\
        &  Steg-PADs & \underline{72.925} 	& \underline{98.975} 	& \underline{97.700} 	& 58.905 	& \underline{98.485} 	& \underline{98.480} 	& \underline{98.085} 	& \underline{89.079}  \\ 
        &  SID~\cite{SID} & 68.159 	& 69.813 	& 61.320 	& 60.722 	& 74.468 	& 74.454 	& 67.213 	& 68.021 \\
        &  SID-PADs & 68.365 	& 62.450 	& 61.145 	& 56.790 	& 70.075 	& 70.080 	& 63.470 	& 64.625  \\
        & SA~\cite{txt_advdetection} & 65.862 	& 72.558 	& 63.304 	& \underline{67.745} 	& 72.087 	& 72.183 	& 73.811 	& 69.650 \\
        & SA-PADs & 65.060 & 69.765 & 59.365 & 62.605 & 66.795 & 66.955 & 67.385 & 65.419 \\
        & AED-PADA & \textbf{73.500} 	& \textbf{99.220} 	& \textbf{98.995} 	& \textbf{68.990} 	& \textbf{99.295} 	& \textbf{99.310} 	& \textbf{99.220} 	& \textbf{91.219}  \\[1.0pt]
        \hline
    \end{tabular}}

    \label{table:sota_svhn}
\end{table*}

\begin{table*}[!t]
    \centering
    \caption{Comparison of the generalization performances on ImageNet between the state-of-the-art adversarial detection methods and our AED-PADA. Averaged Accuracy is the average result across seven unseen state-of-the-art testing adversarial attacks. The bolded and the underlined values represent the best and the second best results for each column, respectively.}

    \resizebox{.85\textwidth}{!}{\begin{tabular}{cccccccccc}
        \hline
        \rule{0pt}{10pt}
        \multirow{2}{*}{\makecell{Dataset \\ (Backbone)}}  & \multirow{2}{*}{Detector} & \multicolumn{7}{c}{Accuracy on Unseen SOTA Adversarial Attacks (\%)} & \multirow{2}{*}{\makecell{Averaged \\ Accuracy~(\%)}} \\
        \cline{3-9}
        \rule{0pt}{9pt}
         & &   APGD & \text{ILA-DA} & Jitter & SSA & TI-FGSM & VMI-FGSM & VNI-FGSM   \\
        \hline
        \rule{0pt}{9pt}
        \multirow{9}{*}{\makecell{ImageNet \\ \\ (ResNet-18)}}  & LID~\cite{lid} &52.868 	&54.078 	&50.991 	&53.263 	&53.793 	&56.566 	&56.549 	&54.015 \\
        &  LID-PADs &57.340 	&63.910 	&55.195 	&58.150 	&58.890 	&60.900 	&61.245 	&59.376  \\
        &  MD~\cite{md} &53.760 	&52.031 	&51.728 	&51.603 	&53.533 	&59.438 	&58.536 	&54.376 \\
        &  MD-PADs &54.560 	&50.425 	&50.685 	&52.765 	&54.695 	&62.165 	&61.040 	&55.191 \\
        &  Steg~\cite{steg} &78.415 	&86.954 	&86.553 	&85.431 	&65.101 	&94.679 	&94.606 	&84.534 \\
        &  Steg-PADs &\underline{90.725} 	&\underline{94.735} 	&\underline{94.700} 	&\underline{94.535} 	&\underline{89.660} 	&\underline{94.875} 	&\underline{94.875} 	&\underline{93.444} \\
        &  SID~\cite{SID} &51.885 	&56.475 	&51.745 	&51.425 	&55.285 	&55.335 	&54.995 	&53.878 \\
        &  SID-PADs &55.070 	&56.975 	&53.785 	&52.335 	&59.790 	&59.860 	&58.400 	&56.602  \\
        &  SA~\cite{txt_advdetection} &85.738 	&87.149 	&85.912 	&85.712 	&86.960 	&85.988 	&86.014 	&86.210 \\
        & SA-PADs & 66.955 &69.230 	&67.665 	&66.930 	&67.385 	&68.490 	&68.550 	&67.886 \\
        & AED-PADA & \textbf{98.830} 	&\textbf{99.935} 	&\textbf{99.965} 	&\textbf{99.965} 	&\textbf{99.070} 	&\textbf{99.965} 	&\textbf{99.960} 	&\textbf{99.670} \\[1.3pt]
        \hline
        \rule{0pt}{9pt}
        \multirow{9}{*}{\makecell{ImageNet \\ \\ (VGG-16)}}  & LID~\cite{lid} & 53.144 	&53.672 	&52.029 	&51.694 	&52.056 	&57.081 	&56.173 	&53.693 \\
        &  LID-PADs &56.295 	&59.330 	&55.110 	&54.825 	&55.695 	&59.960 	&59.310 	&57.218  \\
        &  MD~\cite{md} &53.760 	&52.031 	&51.728 	&51.603 	&53.533 	&59.438 	&58.536 	&54.376 \\
        &  MD-PADs &53.470 	&50.705 	&51.960 	&50.970 	&52.430 	&57.385 	&56.585 	&53.358 \\
        &  Steg~\cite{steg} &78.931 	&90.915 	&85.852 	&84.279 	&66.300 	&94.695 	&94.969 	&85.134 \\
        &  Steg-PADs &90.585 	&94.185 	&\underline{94.115} 	&\underline{94.165} 	&90.375 	&\underline{94.530} 	&\underline{94.535} 	&\underline{93.213} \\
        &  SID~\cite{SID} &41.718 	&50.624 	&50.477 	&44.102 	&51.015 	&51.567 	&51.528 	&48.719 \\
        &  SID-PADs &43.238 	&46.165 	&53.794 	&50.002 	&49.125 	&53.257 	&46.382 	&48.852 \\
        & SA~\cite{txt_advdetection} &87.747 	&93.025 	&89.939 	&88.176 	&86.256 	&92.133 	&92.255 	&89.933 \\ 
        & SA-PADs &\underline{91.590} 	&\underline{94.910} 	&93.415 	&92.310 	&\underline{91.280} 	&94.250 	&94.280 	&93.148 \\
        & AED-PADA &\textbf{98.960} 	&\textbf{99.930} 	&\textbf{99.935} 	&\textbf{99.890} 	&\textbf{98.225} 	&\textbf{99.935} 	&\textbf{99.930} 	&\textbf{99.544} \\[1.3pt]
        \hline
        
    \end{tabular}}

    \label{table:sota_imagenet}
\end{table*}

\subsubsection{Implementation details}
In this paper, adversarial examples are generated by the untargeted white-box attacks with the $l_\infty$ norm constraint. The perturbation of the training adversarial examples is constrained to a challenging scenario, where the maximum magnitude of the adversarial perturbation is set to 2. The step size and number of iterations for adversarial attacks are set to 1/255 and 10, respectively. During PADA stage,  we utilize MFSAN~\cite{aaai-mda} as the basic MUDA method and employ the same training strategy as it. All the adversarial example detection methods are trained consistently for 100 epochs. To evaluate the performance of our proposed method, the widely used Accuracy is employed as the metric for the adversarial example detection task. We utilize MFSAN as the basic MUDA method in the PADA stage of our framework and set the trade-off parameters $\lambda=\gamma=1.0$, which respectively control the importance of $L_d$ and $L_{disc}$. The experiments are conducted on an NVIDIA GeForce RTX 3080Ti GPU.

\subsection{Performance Evaluations}
Here, we compare our method with 5 SOTA detection methods, including LID~\cite{lid}, MD~\cite{md}, Steg~\cite{steg}, SID~\cite{SID} and SA~\cite{txt_advdetection}.

\subsubsection{Generalization performances against unseen adversarial attacks}

Tables \ref{table:sota}, \ref{table:sota_svhn}, and \ref{table:sota_imagenet} present the cross-attack detection results across 7 unseen testing adversarial attacks on CIFAR-10, SVHN, and ImageNet, respectively. Existing adversarial detection methods, which named without `-PADs’, are trained on a single attack with their original settings. Each of their results is the averaged detection accuracy with 10 training attacks being utilized one after another in the training process. On the contrary, those with `-PADs’ are trained on multiple attacks to be fairly compared to our AED-PADA.

These results provide compelling evidence that our approach obviously outperforms these state-of-the-art adversarial detection methods on the generalization ability. Note that the performance of our approach (99.544\%) surpasses that of SID (48.719\%) by a significant margin, i.e., 50.825\%, under the settings of ImageNet and VGG-16. Besides, our superiority is particularly achieved in challenging scenarios, where the maximum magnitude of the adversarial perturbation is merely set to 2. However, in previous studies, the maximum magnitude is typically set to 4 or 8. The reduction in the magnitude of perturbations
significantly increases the difficulty of detecting adversarial examples and reduces the detection performance. The experimental results in Tables \ref{table:sota}, \ref{table:sota_svhn}, and \ref{table:sota_imagenet} demonstrate that our AED-PADA is effective in the scenarios characterized by subtle adversarial perturbations.

As can be observed, our AED-PADA demonstrates better generalization ability on ImageNet compared to that on CIFAR-10 and SVHN, because the image resolution in ImageNet ($224 \times 224$) is higher than that in CIFAR-10 and SVHN ($32 \times 32$). Intuitively, a larger image tends to provide a larger space for adversarial perturbation generation, which makes the differences between the perturbations obtained from various attack methods more pronounced, i.e., reducing the overlaps between different adversarial attacks in the adversarial feature space. By selecting PADs, our AED-PADA is able to occupy a larger feature space, thus achieving a better generalization ability.

Besides, PADs may not be suitable to be directly applied to the existing methods. As can be observed, if the SOTA methods directly adopt PADs for training (denoted as X-PADs), their performances may not always increase. For instance, the generalization performance of LID-PADs on CIFAR-10 is inferior to that of LID. We postulate that this discrepancy can be attributed partially to the reduction in the volume of the training data, and partially to the incompatibility between PADs and the framework of LID. This observation further verifies the effectiveness of our PADA in our AED-PADA framework.

\subsubsection{Generalization performances across different backbones and datasets}
To thoroughly evaluate the generalization ability of adversarial detection methods, it is also vital to assess their performances in the scenarios where the training and testing adversarial examples are from different backbones and datasets. A detection method with better cross-attack, cross-backbone, and cross-dataset capabilities tends to be better suited for practical applications in complex environments.

Table~\ref{table:across} displays the generalization performances of different detection methods across different backbones and datasets. The cross-backbone results are obtained on CIFAR-10, and the cross-dataset results are obtained when the backbone is ResNet-18. In general, results from both scenarios demonstrate that our method exhibits the best generalization ability. Since the structures tend to vary significantly among different types of backbones, which will result in different feature dimensions between training and testing, both LID and MD are not applicable in cross-backbone experiment, i.e., they lack the capability to generalize across backbones, because they utilize the intermediate features of the backbones for detection.

\begin{table*}[!t]
    \centering

    \caption{Comparison of generalization performances across different backbones and datasets. The bolded and the underlined values represent the best and the second best results for each column, respectively.}
    \label{table:across}
    
    \resizebox{.55\columnwidth}{!}{\begin{tabular}{cccccc}

        \hline
        \multirow{4}{*}{Detector} & \multicolumn{5}{c}{Averaged Accuracy~(\%)} \\
        \cline{2-6}
        \rule{0pt}{8pt}
        & \multicolumn{2}{c}{Cross-Backbone} &&\multicolumn{2}{c}{Cross-Dataset } \\
        \cline{2-3} \cline{5-6} 
        \rule{0pt}{13pt}
        & {\makecell{ResNet-18$\rightarrow$ \\VGG-16}} & {\makecell{VGG-16$\rightarrow$ \\ ResNet-18}} && {\makecell{CIFAR-10$\rightarrow$ \\ SVHN}}  & {\makecell{SVHN$\rightarrow$ \\ CIFAR-10}} \\  
        \hline
        \rule{0pt}{9pt}
        LID~\cite{lid} & - & - && \underline{71.619} &\underline{80.233} \\
        MD~\cite{md} & - & - && 62.437 & 61.429 \\
        Steg~\cite{steg} &\underline{82.894} 	&85.157 	&&53.249 	&78.155 \\
        SID~\cite{SID}	&74.169 	&76.458 	&&68.928 	&73.442 \\ 
        SA~\cite{txt_advdetection}	&80.751 	&\underline{87.773} 	&&58.910 	&51.694 \\
        Ours	&\textbf{89.301} 	&\textbf{91.624} 	&&\textbf{82.964} 	&\textbf{85.113} \\ 
        \hline
        
    \end{tabular}}    
    
\end{table*}

\begin{table}[ht]
    \centering
    \begin{minipage}{0.48\textwidth}
    \centering
    \caption{Comparison of generalization performances across different maximum perturbation magnitudes on CIFAR-10 when the backbone is ResNet-18. The maximum magnitude of the testing perturbation is set to 1, 2, 4, 8, respectively. }
    \label{table:eps_cifar10}
    \resizebox{\textwidth}{!}{
    \begin{tabular}{ccccccc}
    \hline
    \multirow{2}{*}{Detector} &Known & &\multicolumn{4}{c}{Unseen} \\
    \cline{2-2} \cline{4-7}
    &2 & &1 &4 &8 &avg \\
    \hline
    LID~[13]	&84.4 &	&56.2 	&71.0 	&76.1 	&67.8 	 \\
    MD~[14]	&69.6 &	&57.1 	&66.2 	&66.4 	&63.2 	\\
    Steg~[15]	&85.2 &	&67.8 	&90.0 	&90.4 	&82.7  \\
    SID~[16]	&76.0 &	&55.2 	&68.0 	&64.6 	&62.6 \\
    SA~[17]	&87.9 &	&75.7 	&89.8 	&88.9 	&84.8 \\
    Ours	&\textbf{94.7} &	&\textbf{89.9} &\textbf{97.0} 	&\textbf{95.2}	&\textbf{94.0} 	 \\
    \hline 
    \end{tabular}}
        
    \end{minipage}
    \hfill
    \begin{minipage}{0.48\textwidth}
    \centering
    \caption{Comparison of generalization performances across different maximum perturbation magnitudes on SVHN when the backbone is ResNet-18. The maximum magnitude of the testing perturbation is set to 1, 2, 4, 8, respectively.}
    \label{table:eps_svhn}
    \resizebox{\textwidth}{!}{

    \begin{tabular}{ccccccc}
    \hline
    \multirow{2}{*}{Detector} &Known &  &\multicolumn{4}{c}{Unseen} \\
    
    \cline{2-2} \cline{4-7}
    &2 & &1 &4 &8 &avg \\
    \hline
    LID~[13]	&69.4 & 	&55.0	&69.9 	&75.0 	&66.6 \\
    MD~[14]	&63.1 	& &55.4 	&63.4 	&65.2 	&61.3 \\
    Steg~[15]	&81.0	& &58.3  	&94.7 	&95.4 	&82.8 \\
    SID~[16]	&70.5 	& &51.9 	&64.8 	&65.4 	&60.7 \\
    SA~[17]	&74.8 	& &55.7 	&83.9 	&84.1 	&74.6 	 \\
    Ours	&\textbf{93.3} &
 &\textbf{89.1} 		&\textbf{98.6} 	&\textbf{99.5} 	&\textbf{95.7} \\

    \hline 
    \end{tabular}}   
    \end{minipage}
\end{table}

\subsubsection{Generalization performances across different maximum perturbation magnitudes}

Here, we evaluate the robustness of the detection models by focusing on a challenging scenario where the maximum perturbation magnitude of the adversarial examples to be detected is unknown. All the detection models are trained on adversarial examples with a maximum perturbation magnitude of 2, while the maximum magnitude of the testing adversarial perturbations is varied across 1, 2, 4 and 8, respectively. Tables~\ref{table:eps_cifar10} and~\ref{table:eps_svhn} present the generalization performances across different maximum perturbation magnitudes on CIFAR-10 and SVHN, respectively. Based on these results, our AED-PADA consistently outperforms the state-of-the-art adversarial detection methods across all the testing scenarios, demonstrating superior generalization ability of our method against the adversarial perturbations with unseen perturbation magnitudes.

\subsection{Effectiveness of AD clustering and PADs Selection}

Selecting multiple adversarial attacks from the same cluster leads to redundancy, as these candidate attacks are quite similar. The combination of them covers a small feature space, causing poor generalization. On the other hand, choosing from different clusters effectively avoids this issue. Consequently, as shown in Table \ref{table:different_souce} and Table~\ref{table:different_souce_v2}, we set up two groups of experiments, i.e., selecting ADs from Same Cluster and Cross Cluster, to verify the effectiveness of the AD Clustering and PADs Selection of AED-PADA, against the 7 unseen attacks. `Same Cluster' and `Cross Cluster' respectively represent that the ADs employed as the sources domains are selected from the same AD cluster and different AD clusters.

Table \ref{table:different_souce} presents the generalization performances on CIFAR-10 and ResNet-18. The result of AD clustering is \{FGSM, PGD, DIM, MI-FGSM, SI-NI-FGSM\}-\{BIM, ILA, YA-ILA\}-\{C\&W, DeepFool\}. For `Same Cluster', the best, worst and mean values of  averaged results are 90.444\%, 93.494\%, and 92.150\%, respectively. For `Cross Cluster', the best, worst, and mean values are 94.694\%, 92.429\%, and 93.892\%, respectively. Similarly, Table~\ref{table:different_souce_v2} shows the results on CIFAR-10 and VGG-16. The result of AD clustering is \{BIM, ILA, YA-ILA, PGD\}-\{DIM, MI-FGSM, FGSM\}-\{CW, DeepFool, SI-NI-FGSM\}. For `Same Cluster', the best, worst and mean values of the averaged results are 87.132\%, 88.019\%, and 87.461\%, respectively. For `Cross Cluster', the best, worst, and mean values of the averaged values are 88.264\%, 87.406\%, and 87.818\%, respectively.

\begin{table}[!t]
    \centering
    \caption{Detection Results of different source domains on CIFAR-10 when the backbone is ResNet-18.}
    
    \resizebox{0.55\columnwidth}{!}{\begin{tabular}{cccc}
        \hline
        Origin & Detector & \makecell{Averaged \\Accuracy~(\%)} & CEFS($\uparrow$) \\
        \hline
        \rule{0pt}{10pt}
        \multirow{3}{*}{\makecell{Same \\ Cluster}} & $\text{FGSM+SI-NI-FGSM+PGD}$ & 90.444 & - \\
        & BIM+ILA+YA-ILA & 92.513 	& - \\
        & MI-FGSM+SI-NI-FGSM+DIM & \textbf{93.494} & - \\
        \hline
        \rule{0pt}{10pt}
        \multirow{8}{*}{\makecell{Cross \\ Cluster}} & DIM+BIM+C\&W & \textbf{94.694} 	& \textbf{80.856} \\
        & MI-FGSM+BIM+C\&W  & 94.621 	& 72.625 \\
        & SI-NI-FGSM+BIM+C\&W & 94.624 	& 58.649 \\
        & FGSM+BIM+C\&W	 & 94.600 	& 58.543 \\
        & PGD+BIM+C\&W	 & 94.222 	& 50.669  \\
        & FGSM+BIM+DeepFool & 93.407 	& 50.142 \\
        & FGSM+ILA+DeepFool	& 92.539 	& 49.818 \\
        & PGD+ILA+DeepFool	& 92.429 	& 48.015 \\
        \hline

    \end{tabular}}

    \label{table:different_souce}
\end{table}

\begin{table}[t]
    \centering
    \caption{Detection Results of different source domains on CIFAR-10 when the backbone is VGG-16.}
    \resizebox{.55\columnwidth}{!}{\begin{tabular}{cccc}
        \hline
        Origin & Detector & \makecell{Averaged \\Accuracy~(\%)} & CEFS($\uparrow$) \\
     
        \hline
        \rule{0pt}{9pt}
        \multirow{6}{*}{\makecell{Same \\ Cluster}} & BIM+ILA+YA-ILA & 87.389 & - \\
        & BIM+ILA+PGD & 87.153 & - \\
        & BIM+YA-ILA+PGD & 87.132 & - \\
        & ILA+YA-ILA+PGD  & 87.439 & - \\
        & DIM+MIM+FGSM  & 87.634 & - \\
        & $\text{CW+DeepFool+SI-NI-FGSM}$ & \textbf{88.019} & - \\

        \hline
        \rule{0pt}{9pt}
        \multirow{8}{*}{\makecell{Cross \\ Cluster}} & 
        ILA+DIM+CW	 & \textbf{88.264} 	& \textbf{18.692} \\
        & PGD+DIM+CW	 & 88.236 	& 18.605 \\
        & ILA+FGSM+CW	 & 87.969 	& 18.532 \\
        & YA-ILA+FGSM+CW	 & 87.897 	& 18.527 \\
        & YA-ILA+MI-FGSM+SI-NI-FGSM	 & 87.670 	& 18.518 \\
        & BIM+DIM+SI-NI-FGSM	 & 87.676 	& 18.410 \\
        & YA-ILA+FGSM+SI-NI-FGSM	 & 87.425 	& 18.388 \\
        & YA-ILA+DIM+SI-NI-FGSM	 & 87.406 	& 18.385 \\
        \hline
    \end{tabular}}

    \label{table:different_souce_v2}    
\end{table}

Based on the results, we can obtain three observations. Firstly, the mean results of `Cross Cluster' selection is higher than of `Same Cluster' selection, and the best result of `Cross Cluster' selection is also superior. Specially, On the setting of CIFAR-10 and ResNet-18, the mean results of `Cross Cluster' selection is even higher than of the best results of `Same Cluster' selection. This verifies the effectiveness of our AD Clustering. Apparently, source domains from `Cross Cluster' can certainly enhance the generalization ability of the detection methods. 

Secondly, although `Cross Cluster' selection in general achieves better results, it cannot guarantee that the randomly selected ADs (from `Cross Cluster's) always give better performance than the results of `Same Cluster' selected ADs. When the PADs with a very low CEFS score are employed as the source domains for detection, the generalization performance may be worse than that of `Same Cluster' selection. This observation further verifies the effectiveness of our PADs Selection. 

Thirdly, as shown in the `Cross Cluster' part of table~\ref{table:different_souce} and table~\ref{table:different_souce_v2}, for the majority of results, a higher CEFS value of PADs induces a higher averaged accuracy of the proposed detection. This observation suggests that the proposed CEFS is a proper guide for selecting PADs, i.e., a higher CEFS score indicates that the corresponding PADs can cover a larger proportion of the entire feature space, and the PADA stage, which uses the PADs as source domains, gives a better generalization performance.

\subsection{The effect of different numbers of clusters}

Here, we apply AD Clustering to 10 adversarial attack methods in the training set and present the impact of different cluster numbers $K$ (the number of adversarial attacks in PADs). As depicted in Table \ref{tab:cluster_count}, the detection performance increase when $K$ increases, until a certain value of $K$ is achieved. This trend is attributed to the utilization of an increased number of ADs as the source domains, which undeniably provides larger coverage of the entire feature space. Note that the values with $\dag$ represent the generalization performance of our AED-PADA, in which the number of clusters is automatically determined based on the CH score, and the bolded values represent the best generalization performance. 

As can be observed, our automatic CH score based method can yield superior performance on SVHN with VGG-16 being the backbone, and give comparable performance to the best result on other settings. Although the best result with manually selecting $K$ exceeds our automatic method for 0.2\%-0.3\% in terms of the averaged accuracy, it requires $2.3\times$ more parameters and gives a $3\times$ slower speed. Moreover, in real-world scenarios, the adversarial attacks in the training set may far exceed 10 types, and it is clearly unwise to test the detection performance of each clustering result individually. Consequently, this suggests that our method actually achieves a better balance between the training costs and generalization performances.

\begin{table*}[t]
    \centering
    \caption{Detection Performance across seven unseen testing adversarial attacks with different number of clusters $K$.}
    \resizebox{.65\columnwidth}{!}{\begin{tabular}{ccccccc}
        \hline
        \rule{0pt}{8pt}
        \multirow{2}{*}{\makecell{Dataset \\ \rule{0pt}{8pt} (Backbone)}} & \multicolumn{6}{c}{Averaged Accuracy (\%)} \\
        \cline{2-7}
        \rule{0pt}{10pt}
        &  $K=2$ & $K=3$ & $K=4$ & $K=5$ & $K=6$ & $K=7$ \\
        \hline
        \rule{0pt}{17pt}
        \makecell{CIFAR-10  \\ \rule{0pt}{7.5pt} (ResNet-18)} & 91.936 &$94.694^{\dag}$ & 94.784 & 94.571 & 94.561 & \textbf{95.018} \\
        \hline
        \rule{0pt}{17pt}
        \makecell{CIFAR-10  \\ \rule{0pt}{7.5pt} (VGG-16)} & 87.946 &$88.264^{\dag}$ & 87.964 & 87.991 & 88.104 & \textbf{88.498} \\
        \hline
        \rule{0pt}{17pt}
        \makecell{SVHN  \\ \rule{0pt}{7.5pt} (ResNet-18)} & 91.497 &$94.344^{\dag}$ &91.257 & 91.726 & \textbf{94.999} & 94.801 \\
        \hline
        \rule{0pt}{17pt}
        \makecell{SVHN  \\ \rule{0pt}{7.5pt} (VGG-16)} & 89.183 & 89.269 &$\textbf{91.219}^{\dag}$ & 90.056 & 90.026 & 90.713 \\
        \hline
    \end{tabular}}

    \label{tab:cluster_count}
\end{table*}

\begin{table}[t]
    \centering
    \caption{Detection Performance of different feature extractions in PADA of our AED-PADA.}
    \resizebox{.65\columnwidth}{!}{\begin{tabular}{ccccc}

        \hline
        \rule{0pt}{10pt}
        \multirow{2}{*}{Dataset} &\multirow{2}{*}{\makecell{Feature \\ Extraction}}  & \multicolumn{2}{c}{Averaged Accuracy (\%)} & \multirow{2}{*}{Average} \\
        \cline{3-4}
        \rule{0pt}{9pt}
        & & ResNet-18 & VGG-16 & \\
        \hline
        \rule{0pt}{10pt}
        \multirow{3}{*}{CIFAR-10} & spatial & 93.411 & 88.251 & 90.831 \\
        & freq & 93.379 & 86.755 & 90.067 \\
        & spatial + freq & \textbf{94.694} & \textbf{88.264} & \textbf{91.479} \\

        \hline
        \rule{0pt}{10pt}
        \multirow{3}{*}{SVHN} & spatial & \textbf{94.612} & 89.721 & 92.166 \\
        & freq & 88.157 & \textbf{91.580} & 89.869 \\
        & spatial + freq & 94.344 & 91.219 & \textbf{92.781} \\

        \hline         
    \end{tabular}}

    \label{table:common_extractor}    
\end{table}

\subsection{Effectiveness of the adversarial feature enhancement in PADA}
In this experiment, we utilize three distinct feature extractions, i.e., spatial feature extraction, frequency feature extraction, and the adversarial feature enhancement (spatial + freq) in our framework. Still 7 test unseen attacks are employed and the averaged results are reported in Table \ref{table:common_extractor}. As can be observed, in both datasets, the average results of our adversarial feature enhancement(spatial + freq) exhibit superior performances than both the spatial feature extraction and frequency feature extraction. This indicates that using adversarial feature enhancement as the feature extraction method for PADA is effective. Furthermore, whether it is spatial feature extraction or frequency feature extraction, the performance of single domain feature extraction across different backbones is inconsistent, as evidenced by the fluctuated averaged detection performance across different backbones on SVHN. Our adversarial feature enhancement, which combines both spatial and frequency aspects, can more comprehensively capture adversarial perturbation signals, yielding superior adversarial detection performance.

\subsection{The compatibility with different MUDA methods}

To demonstrate the compatibility of our framework with the existing Multi-source Unsupervised Domain Adaptation~(MUDA) methods, we employ four widely used Multi-source Unsupervised Domain Adaptation~(MUDA) methods, M$^3$DA~\cite{moment-mda}, DARN~\cite{DARN}, MDMN~\cite{mdmn} and MFSAN~\cite{aaai-mda}. All four MUDA methods have utilized the adversarial feature enhancement as the feature extraction component. Table \ref{table:different_MDA} presents that the performances of employing all four MUDA methods can surpass the existing SOTA adversarial detection methods. This verifies that our framework possesses excellent compatibility with existing MUDA methods. In other words, The first exploration of MUDA methods in adversarial example detection has proven to be successful, and Principal Adversarial Domain Adaptation can effectively transfer the knowledge from PADs to the unseen target domain. Besides, since the average result of MFSAN outperforms other MUDA methods, we select MFSAN as the basic MUDA component in our PADA.

\begin{table}[t]
    \centering
    \caption{Detection performance of our framework with different MUDA methods.}
    \resizebox{.75\columnwidth}{!}{\begin{tabular}{ccccccc}
        \hline
        \rule{0pt}{10pt}
        \multirow{2}{*}{Dataset} &\multirow{2}{*}{Backbone} &\multicolumn{4}{c}{Averaged Accuracy of different MUDA methods~(\%)} \\
        \cline{3-6}
        \rule{0pt}{10pt}
        & & M$^3$DA~\cite{moment-mda} &DARN~\cite{DARN} &MDMN~\cite{mdmn} &MFSAN (Ours)~\cite{aaai-mda} \\

        \hline
        \rule{0pt}{9pt}
        CIFAR-10 &ResNet-18 &\textbf{94.976} &93.948 &93.832 &94.694 \\
        CIFAR-10 &VGG-16 &87.819 &87.624 &\textbf{91.463} &88.864\\
        SVHN &ResNet-18 &93.892 &93.934 &93.888 &\textbf{94.344}\\
        SVHN &VGG-16 &90.960 &\textbf{92.244} &89.826 &91.219 \\
        \hline
        \multicolumn{2}{c}{Average} &91.912 &91.938 &92.252 &\textbf{92.280} \\
        \hline     
    \end{tabular}}
    
    \label{table:different_MDA}
\end{table}

\subsection{Parameters sensitivity.} We utilize MFSAN as the basic MUDA method in the PADA stage of our framework, and $\lambda=\gamma=1.0$ control the importance of $L_d$ and $L_{disc}$, respectively. To study the parameters sensitivity, we sample the values of $\lambda$ and $\gamma$ from \{0.5, 1.0, 2.0\}, and perform the experiments under the settings of CIFAR-10 and ResNet-18.

\begin{table}[t]
    \small
    \centering
    \caption{Sensitivity of the parameters $\lambda$ and $\gamma$ in the PADA stage.  
    }
    \scriptsize
    \resizebox{0.40\columnwidth}{!}{\begin{tabular}{ccc}
        \hline
        $\lambda$ & $\gamma$ & Averaged Accuracy~(\%) \\
        \hline
        \rule{0pt}{7pt}
        $\lambda$=0.5 & $\gamma$=0.5 & 94.503 \\
        $\lambda$=0.5 & $\gamma$=1.0 & 94.591 \\
        $\lambda$=0.5 & $\gamma$=2.0 & 94.355 \\
        $\lambda$=1.0 & $\gamma$=0.5 & 94.646 \\
        $\lambda$=1.0 & $\gamma$=1.0 & \textbf{94.694} \\
        $\lambda$=1.0 & $\gamma$=2.0 & 94.603 \\
        $\lambda$=2.0 & $\gamma$=0.5 & 94.598 \\
        $\lambda$=2.0 & $\gamma$=1.0 & 94.629 \\
        $\lambda$=2.0 & $\gamma$=2.0 & 94.223 \\
        \hline
    \end{tabular}}

    \label{table:different_gamma}
\end{table}

Table~\ref{table:different_gamma} indicates that our proposed method maintains high performance consistency under various parameter settings when employing MFSAN as the basic MUDA method. The best averaged accuracy is 94.694\% when $\lambda=\gamma=1.0$. Variations of the parameters typically do not induce the fluctuations in the detection performance. The mean value and the standard deviation of all the averaged accuracies are 94.538\% and 0.154\%, respectively. This indicates that our PADA framework is effective and robust, since it shows insensitivity to different parameters and consistently offers high detection accuracy.

\subsection{The computational costs}

\begin{table}[t]
    \large
    \centering
    \caption{Comparison of the time and hardware costs. Batch size for each method is 1000.}

    \resizebox{0.55\columnwidth}{!}{\begin{tabular}{ccccccc}
        \hline
        \rule{0pt}{10pt}
        \multirow{3}{*}{Cost} &\multicolumn{6}{c}{Detector} \\
        \cline{2-7}        
        \rule{0pt}{16pt}
        & \makecell{LID \\ \cite{lid}} & \makecell{MD \\ \cite{md}} & \makecell{Steg \\ \cite{steg}} & \makecell{SID \\ \cite{SID}} & \makecell{SA \\ \cite{txt_advdetection}} & Ours \\
        \hline
        \rule{0pt}{13pt}
        Time (ms/image) & 8.13 & 18.05 & 23.85 & 2.16 &1.21 & \textbf{0.60} \\
        \hline
        \rule{0pt}{13pt}
        Hardware (MB) & 7458 & 7248 & 1614 & \textbf{1592} &8564 & 6124 \\
        \hline
    \end{tabular}}

    \label{table:exp}
\end{table}

Table~\ref{table:exp} presents the time and hardware costs when deploying the state-of-the-art adversarial example detection methods and our AED-PADA. The results indicate that our method has the minimal time expenses and moderate hardware (GPU memory) expenses under the same settings. All experiments are conducted on an NVIDIA GeForce RTX 3080Ti GPU under CIFAR-10 and ResNet-18 settings.

Furthermore, our AED-PADA is capable of deployment in real-world scenarios. Real-world scenarios demand the adversarial example detection methods not only possess the capability for real-time detection but also maintain robust detection performance in the environments where the adversarial attacks, datasets, and backbones during testing are entirely unseen. Firstly, we only train once, ready for real-world deployment without retraining for new unseen attacks. We have the shortest inference time, which is the best real-time detection performance. Secondly, we train from earlier attacks and test on more advanced attacks. Table~\ref{table:sota} demonstrates that under this real-world-aligned setting, our AED-PADA exhibits superior generalization performance, and would maintain considerable detection capabilities against future unseen attacks. Lastly, with its strong performance across various backbones and datasets depicted in the Table~\ref{table:across}, our proposed method performs better in the environments with unseen datasets and backbones. Consequently, our AED-PADA is well-suited for practical applications in the complex real-world environments.

\section{Conclusion}

In this paper, we proposed a novel and effective adversarial example detection method, named Adversarial Example Detection via Principal Adversarial Domain Adaptation, to improve the generalization ability of adversarial detection. Specifically, AED-PADA contains two stages, i.e., Principal Adversarial Domains Identification~(PADI) and Principal Adversarial Domain Adaptation~(PADA). In PADI, we acquired ADs from scratch and constructed PADs as the source domains for PADA. In PADA, we proposed an adversarial feature enhancement based Multi-source Unsupervised Domain Adaptation framework, which is compatible with various existing MUDA methods, to effectively leverage PADs to achieve adversarial example detection. Experimental results demonstrated the superiority of our work, compared to the state-of-the-art detection methods.

\begin{acks}
This work was supported in part by the National Natural Science Foundation of China under Grant 62272020, U20B2069 and 62176253, in part by the State Key Laboratory of Complex \& Critical Software Environment under Grant SKLSDE2023ZX-16, and in part by the Fundamental Research Funds for Central Universities.
\end{acks}

\bibliographystyle{unsrt}

\bibliography{reference}


\end{document}